\PassOptionsToPackage{table,dvipsnames}{xcolor} 
\documentclass[letterpaper, 10pt, conference]{ieeeconf} 
\usepackage{amsmath} 
\usepackage{amssymb}  
\usepackage{graphicx}
\usepackage{url}
\DeclareUnicodeCharacter{03A7}{\ensuremath{\chi}}
\usepackage{hyperref}
\usepackage{graphicx}

\title{\LARGE \bf
Multi-Head Adaptive Graph Convolution Network for Sparse Point Cloud-Based Human Activity Recognition
}

\author{Vincent Gbouna Zakka$^{1*}$ Luis J. Manso $^{2}$ Zhuangzhuang Dai $^{3}$\\
\normalsize $^{1,2,3}$Dept. of Applied AI and Robotics,
        Aston University, Birmingham, United Kingdom\\
{$^{*}$Corresponding address: \tt\small vzakk22@aston.ac.uk}
}

\begin{document}

\maketitle
\thispagestyle{empty}
\pagestyle{empty}

\begin{abstract}
Human activity recognition is increasingly vital for supporting independent living, particularly for the elderly and those in need of assistance. Domestic service robots with monitoring capabilities can enhance safety and provide essential support. Although image-based methods have advanced considerably in the past decade, their adoption remains limited by concerns over privacy and sensitivity to low-light or dark conditions. As an alternative, millimetre-wave (mmWave) radar can produce point cloud data which is privacy-preserving. However, processing the sparse and noisy point clouds remains a long-standing challenge. While graph-based methods and attention mechanisms show promise, they predominantly rely on “fixed” kernels—kernels that are applied uniformly across all neighbourhoods—highlighting the need for adaptive approaches that can dynamically adjust their kernels to the specific geometry of each local neighbourhood in point cloud data. To overcome this limitation, we introduce an adaptive approach within the graph convolutional framework. Instead of a single shared weight function, our Multi-Head Adaptive Kernel (MAK) module generates multiple dynamic kernels, each capturing different aspects of the local feature space. By progressively refining local features while maintaining global spatial context, our method enables convolution kernels to adapt to varying local features. Experimental results on benchmark datasets confirm the effectiveness of our approach, achieving state-of-the-art performance in human activity recognition. Our source code is made publicly available at: \url{https://github.com/Gbouna/MAK-GCN}.
\end{abstract}
\section{INTRODUCTION}
Human activity recognition has become increasingly important in monitoring activities of daily living, health status, and general well-being, especially for the elderly and individuals requiring support to live independently~\cite{GayaMorey2024}. By providing continuous observation and alerts when necessary, such solutions can significantly improve quality of life. With the slow but steadily increasing adoption of domestic service robots~\cite{Peijun}, the ability to monitor activities is a compelling feature. If equipped with monitoring capabilities, these robots could simultaneously deliver services and ensure the safety and support of end users~\cite{Sedaghati2025}.

\begin{figure}[tp]
  \centering
  \includegraphics[scale=0.3]{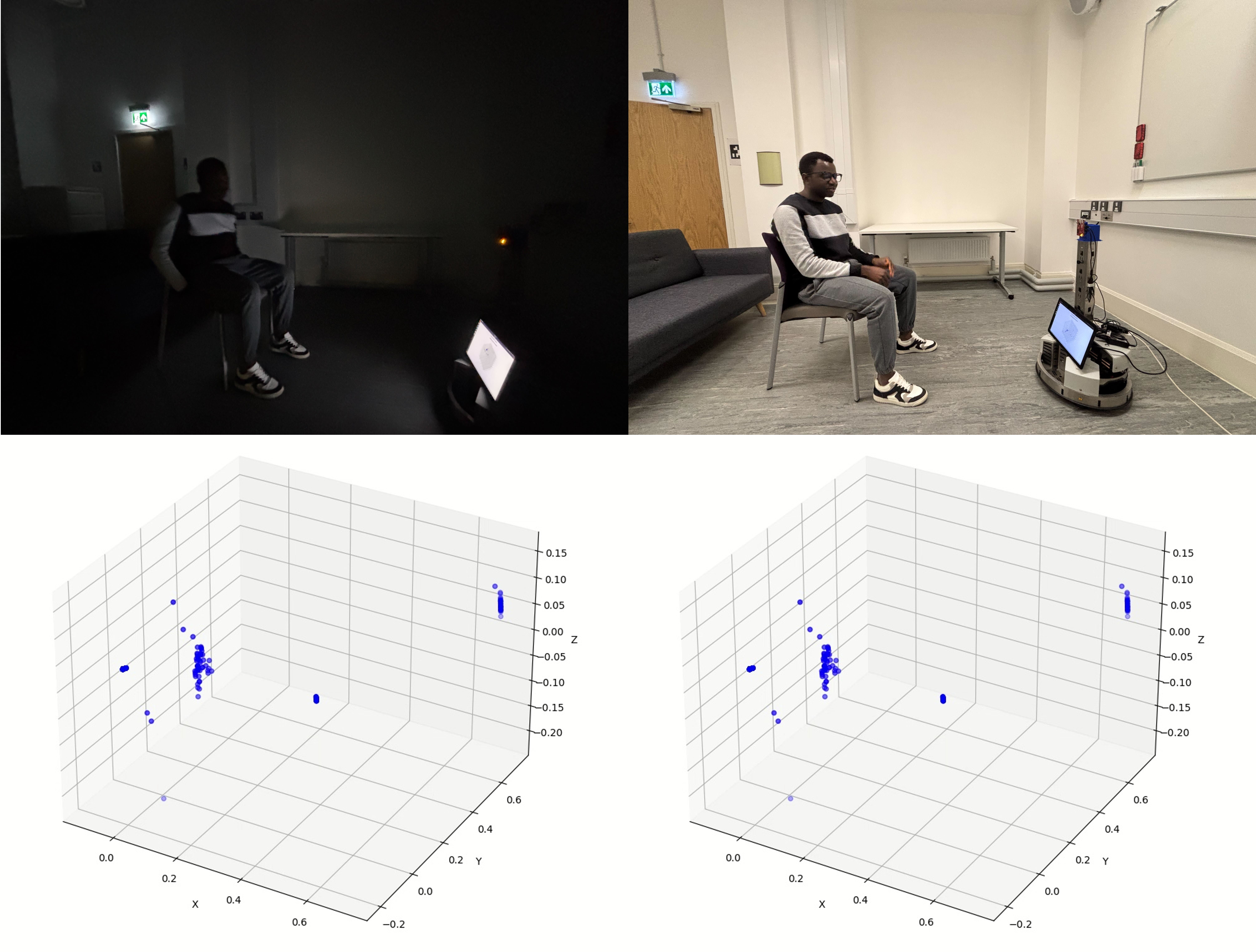}
  \caption{mmWave radar deployed on a Robotino robot for activity monitoring in both dark and illuminated conditions}
  \label{fig:Intro}
\end{figure}

Progress has been made in deploying robots for activity monitoring, with many systems relying on RGB cameras to capture activities of daily living~\cite{info11020075}. Although camera-based approaches have advanced considerably, they face certain limitations, such as privacy concerns, and poor performance on low lighting conditions~\cite{Bhola2024}.
Recent research exploring sensors that overcome these issues has focused on mmWave radar, which produces point cloud data that is inherently privacy-preserving and functions effectively in diverse settings (see Fig.~\ref{fig:Intro}).
Despite its potential, mmWave radar data remains challenging to process due to its unstructured and sparse nature. Early solutions, such as PointNet~\cite{qi2016pointnet}, paved the way for numerous variations~\cite{Charles}. More recently, graph-based methods have attracted attention~\cite{Wang}. However, these methods commonly rely on the same learned weights for all point pairs, disregarding the variations in their feature correspondences. To address this issue, various strategies have been proposed, inspired by attention mechanisms~\cite{Lei}. Although these methods use attention to adjust feature weighting, their underlying convolution kernels remain essentially fixed, limiting their ability to adapt to the local geometry and capture the most relevant elements in each neighbourhood.
To overcome the limitations of fixed convolution kernels in capturing diverse feature correspondences among points, we propose an adaptive approach that learns distinct kernels for each pair of points. We propose the Multi-Head Adaptive Kernel (MAK) module, which can learn context-specific convolutions. Our experiments show that MAK-GCN networks achieve state-of-the-art performance on two mmWave radar-based human activity recognition benchmarks. MAK generates multiple sets of dynamic kernels --each head capturing a distinct aspect of the local feature space. Combined with the overall network design, which ensures that local features are progressively enhanced while preserving the global spatial context of the point cloud, this work aims to improve the ability of a network to dynamically tailor convolution kernels to diverse feature correspondences, ultimately enabling robust and discriminative feature learning for human activity recognition.
Experimental results on the MMActivity~\cite{RadHAR} and MiliPoint~\cite{Milipoint} point cloud datasets for human activity recognition demonstrate the effectiveness of the proposed method, achieving new state-of-the-art surpassing previous methods significantly.
\section{RELATED WORKS}
To address the irregular nature of point clouds, state-of-the-art methods process raw point cloud data directly rather than relying on intermediate representations~\cite{qi2016pointnet}. Graph-based approaches represent points as nodes in a graph, with edges formed based on spatial or feature relationships~\cite{Wang}. While graphs naturally capture local geometric structures, their irregularity makes them difficult to process. Several studies have leveraged the graph-based approach to extract local geometric features. For example,~\cite{Wang} selects nearest neighbours in feature space and applies EdgeConv for feature extraction, and~\cite{Federico} models convolution using Gaussian mixture models within a local pseudo-coordinate system. To enhance performance, various approaches~\cite{Lei} incorporate learned feature-based weights. Nevertheless, they continue to rely on fixed convolution kernels, restricting their ability to flexibly adapt to each local neighbourhood and emphasise the most relevant features. To overcome this limitation, we introduce the Multi-Head Adaptive Kernel (MAK) module, which generates multiple sets of dynamic kernels—each head capturing a distinct aspect of the local feature space.
\section{METHOD}
\begin{figure*}[thpb]
  \centering
  \includegraphics[scale=0.35]{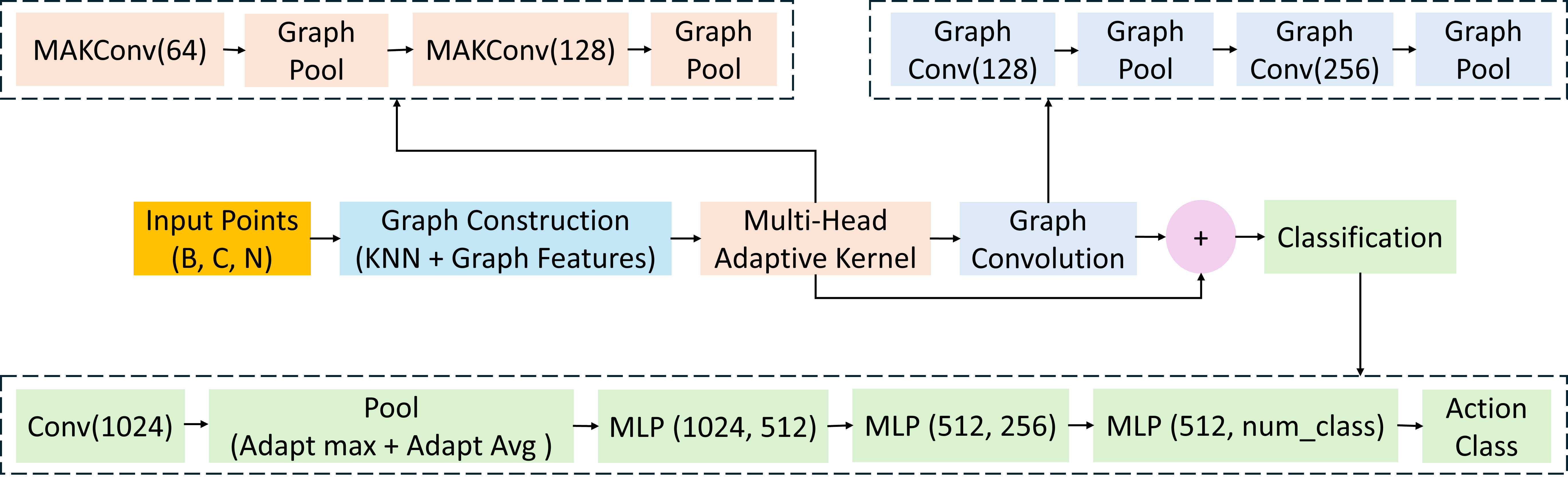}
  \caption{Architecture of the proposed MAKGConvNet: A graph is constructed from input data for convolution via the MAK module, followed by a common graph convolution. The outputs are concatenated and passed to the classification module.}
  \label{fig:model}
\end{figure*}

This section details the proposed model architecture, beginning with a discussion of its main components, followed by an overview of the overall network design as presented in Fig.~\ref{fig:model}.
\subsection{Graph Feature Extraction}
First, we construct a graph representation of the point cloud using a K-Nearest Neighbours (KNN) algorithm to identify local neighbourhoods. Let \( X \in \mathbb{R}^{C \times N} \) be the feature matrix for a single sample, where \( C \) denotes the number of channels and \( N \) the number of points. For a mini-batch of \( B \) samples, the tensor is of shape \( (B, C, N) \).
\subsubsection{KNN Function}
The KNN function computes the pairwise squared Euclidean distances between points. Given an input tensor \(X \in \mathbb{R}^{C \times N}\) (or \((B, C, N)\) for a batch of \(B\) samples), the squared distance between two points \(x_i\) and \(x_j\) is defined as follows:
\begin{equation}\label{eq:euclidean_distance}
d(x_i, x_j)^2 = \|x_i - x_j\|^2 = \|x_i\|^2 + \|x_j\|^2 - 2 \langle x_i, x_j \rangle,
\end{equation}
where \(\langle \cdot,\cdot \rangle\) denotes the inner product. In the implementation, the inner product term is computed by the matrix multiplication \(-2\,X^\top X\). Simultaneously, the squared norms \(\|x_i\|^2\) for each point are calculated and arranged so that they can be broadcasted appropriately. The pairwise distance matrix \(D\) is then obtained as follows:
\begin{equation}\label{eq:pairwise_distance}
D = -\text{S} - \Big(-2\,X^\top X\Big) - \text{S}^\top,
\end{equation}
with \(\text{S}\) representing the vector of squared norms. Finally, the indices corresponding to the \(k\) smallest distances—indicative of the nearest neighbours—are selected using a top-\(k\) operation. This yields an index tensor of shape \((B, N, k)\), which effectively captures the neighbourhood structure necessary for subsequent graph-based feature extraction.
\subsubsection{Graph Feature Extraction Function}
The graph feature extraction module constructs a local neighbourhood representation that captures absolute and relative point features. Given an input tensor, each point \( x_i \) is associated with its \( k \) nearest neighbours. Denote by \( \mathcal{N}(i) \) the set of indices corresponding to the \( k \) nearest points to \( x_i \), as determined by the KNN algorithm. Subsequently, the function gathers the features of the \( k \) nearest neighbours for each point and computes the edge features by taking the difference between a neighbour’s feature \( x_j \) and the central point’s feature \( x_i \):
\begin{equation}\label{eq:edge_features}
f_{ij}^{\text{diff}} = x_j - x_i, \quad \forall j \in \mathcal{N}(i),
\end{equation}
This difference emphasises local geometric variations. The function then constructs an augmented feature vector by concatenating these different features with the original point features:
\begin{equation}\label{eq:augmented_feature_vector}
f_i = \Big[ \{f_{ij}^{\text{diff}}\}_{j \in \mathcal{N}(i)}, \, x_i \Big].
\end{equation}
The final output is a tensor of dimensions \( (B, 2C, N, k) \), where the first \( C \) channels represent the relative differences and the remaining \( C \) channels retain the absolute features of each point.
\subsection{Multi-Head Adaptive Kernel}
The Multi-Head Adaptive Kernel (MAK) module is an enhanced adaptive convolution operator that generates dynamic kernels based on the input feature. It comprises a multilayer perceptron, multiple filtering heads, and utilises residual connections to ensure more stable training. The overall process can be described in three main stages: dynamic kernel generation, multi-head filtering, and output integration with residual connections. Initially, given an input feature map \( y \in \mathbb{R}^{\text{feat\_channels} \times N \times k} \), the module generates dynamic kernel weights through a series of convolutional operations. The first stage is a feature transformation defined by
\begin{equation}\label{eq:initial_feature_transformation}
    y_0 = \mathrm{LeakyReLU}\big(\mathrm{BN}_0(\mathrm{Conv}_0(y))\big),
\end{equation}
where $BN_0$ is batch normalisation.
This is followed by a further transformation
\begin{equation}\label{eq:intermediate_feature_transformation}
    y_1 = \mathrm{LeakyReLU}\big(\mathrm{BN}_{\text{mid}}(\mathrm{Conv}_{\text{mid}}(y_0))\big),
\end{equation}
Subsequently, a final convolution is applied to generate the dynamic kernel weights:
\begin{equation}\label{eq:dynamic_kernel_generation}
    W = \mathrm{Conv}_1(y_1),
\end{equation}
where \( W \) has dimensions \( (B, \text{out\_channels} \times \text{in\_channels} \times \text{num\_heads}, N, k) \). This means that for each of the \( N \) points and \( k \) neighbouring positions, the module produces \(\text{num\_heads}\) sets of filters. To separate these heads, \( W \) is reshaped as follows:
\begin{equation}\label{eq:reshaped_dynamic_weights}
    W \in \mathbb{R}^{B \times N \times k \times \text{out\_channels} \times \text{in\_channels} \times \text{num\_heads}}
\end{equation}
This reshaping disentangles the multiple filter sets (heads) so that each head \( h \) (where \( h = 1, 2, \ldots, \text{num\_heads} \)) has its own dynamic kernel 
\begin{equation}\label{eq:reshaped_filters}
    W^{(h)} \in \mathbb{R}^{B \times N \times k \times \text{out\_channels} \times \text{in\_channels}}.
\end{equation}
Simultaneously, the input feature $x \in \mathbb{R}^{\text{in\_channels} \times N \times k}$
is rearranged to align with the dynamic kernels $x' \in \mathbb{R}^{B \times N \times k \times \text{in\_channels} \times 1}$.
For each head \( h \), the corresponding dynamic filter \( W^{(h)} \) is applied to \( x' \) through matrix multiplication:
\begin{equation}\label{eq:head_wise_dynamic_filtering}
    \mathbf{out}_h = W^{(h)} \cdot x'
\end{equation}
where the resulting output for each head is of size \( (B, N, k, \text{out\_channels}) \). The outputs from all heads are then aggregated by summing across the head dimension:
\begin{equation}\label{eq:aggregation_across_heads}
    \mathbf{out} = \sum_{h=1}^{\text{num\_heads}} \mathbf{out}_h,
\end{equation}
After this multi-head filtering, the tensor is permuted back to its original spatial arrangement, yielding a feature map of dimensions \( (B, \text{out\_channels}, N, k) \).
If residual connections are enabled, the processed output is added to the input \( x \). If the number of input channels does not match the number of output channels, a projection is applied to \( x \) through a \( 1 \times 1 \) convolution followed by batch normalisation, ensuring that the dimensions are compatible before addition. This projected identity, denoted as \(\mathbf{I}\), is then added element-wise to the aggregated output. The final output is thus given by
\begin{equation}\label{eq:final_output}
    \mathbf{out}_{\text{final}} = \mathrm{LeakyReLU}\Big(\mathrm{BN}_{\text{out}}\big(\mathbf{out} + \mathbf{I}\big)\Big),
\end{equation}
where \(\mathbf{I}\) represents either the original \( x \) or its projected version.
\subsection{Overall Network Architecture}
The overall network architecture is designed to extract local and global features from point cloud data in a hierarchical manner, progressively extracting features for robust action recognition. The model comprises five stages, each of which leverages graph-based feature extraction and dynamic filtering to capture relationships among points. The five stages are: graph construction, multi-head dynamic filtering, local feature aggregation, feature fusion, and final classification as shown in Fig.~\ref{fig:model}.
\subsubsection{Stage 1: Graph Construction}
Given an input point cloud, a common neighbourhood structure is computed using the graph feature extraction module. This produces a set of adjacency indices \(\mathcal{I}\) shared across layers. Formally, for each point \( x_i \), its \( k \) nearest neighbours are identified such that
\begin{equation}\label{eq:knn_adjacency}
    \mathcal{I}(i) = \{ j \mid d(x_i, x_j) \text{ is among the } k \text{ smallest} \},
\end{equation}
where \( d(\cdot,\cdot) \) is the squared Euclidean distance.
\subsubsection{Stage 2: Multi-Head Dynamic Filtering}
The network uses the MAK modules in the first two layers to extract local features. In the first layer, both the geometric and feature representations are derived from the raw input points. Specifically, two graph features are computed using the same neighbourhood structure \(\mathcal{I}\) obtained through the graph construction module. The first set, denoted as $\text{feat}_x^{(1)} = \text{GraphFeature}(X, k, \mathcal{I})$ captures the intrinsic feature differences, while the second set, $\text{geo}_x^{(1)} = \text{GraphFeature}(X, k, \mathcal{I}),$ encodes the spatial geometry of the points. These similar yet complementary representations are input into the MAK module, yielding
\begin{equation}\label{eq:X1}
    X_1 = \text{MAK}_1\big(\text{geo}_x^{(1)}, \text{feat}_x^{(1)}\big) \in \mathbb{R}^{64 \times N \times k}.
\end{equation}
Subsequently, a max pooling operation is applied along the neighbourhood dimension \(k\) to aggregate the local features:
\begin{equation}\label{eq:x1}
    x_1 = \max_{j=1,\ldots,k} X_1(:, :, j) \in \mathbb{R}^{64 \times N}.
\end{equation}
In the second layer, the network refines the local features by employing an asymmetric strategy. Here, the feature representation is updated using the output \( x_1 \) from the first layer, while the geometric information remains anchored to the original input \( X \). Formally, the feature-based graph features are computed as $\text{feat}_x^{(2)} = \text{GraphFeature}(x_1, k, \mathcal{I})$
whereas the geometric features are still derived from the raw points $\text{geo}_x^{(2)} = \text{GraphFeature}(X, k, \mathcal{I})$
This design choice ensures that while the local features are progressively refined, the global spatial context provided by the original point cloud is preserved. The second MAK module then processes these inputs to produce
\begin{equation}\label{eq:X2}
    X_2 = \text{MAK}_2\big(\text{geo}_x^{(2)}, \text{feat}_x^{(2)}\big) \in \mathbb{R}^{64 \times N \times k},
\end{equation}
which is aggregated via max pooling to obtain the refined local feature map:
\begin{equation}\label{eq:x2}
    x_2 = \max_{j=1,\ldots,k} X_2(:, :, j) \in \mathbb{R}^{64 \times N}.
\end{equation}
\subsubsection{Stage 3: Deeper Convolutional Feature Extraction}
Subsequent layers employ conventional convolutional modules applied to graph features computed from the previously obtained local representations. In the third layer, graph features derived from \( x_2 \) are processed by a convolutional block:
\begin{equation}\label{eq:third_layer_convolution}
    X_3 = \mathrm{Conv3}\Big(\text{GraphFeature}(x_2, k, \mathcal{I})\Big) \in \mathbb{R}^{128 \times N \times k},
\end{equation}
with max pooling yielding
\begin{equation}\label{eq:third_layer_aggregation}
    x_3 = \max_{j=1,\ldots,k} X_3(:, :, j) \in \mathbb{R}^{128 \times N}.
\end{equation}
Similarly, in the fourth layer, graph features from \( x_3 \) are fed into another convolutional block:
\begin{equation}\label{eq:fourth_layer_convolution}
    X_4 = \mathrm{Conv4}\Big(\text{GraphFeature}(x_3, k, \mathcal{I})\Big) \in \mathbb{R}^{256 \times N \times k},
\end{equation}
followed by max pooling:
\begin{equation}\label{eq:fourth_layer_aggregation}
    x_4 = \max_{j=1,\ldots,k} X_4(:, :, j) \in \mathbb{R}^{256 \times N}.
\end{equation}
\subsubsection{Stage 4: Feature Fusion and Global Descriptor}
The local features \( x_1 \), \( x_2 \), \( x_3 \), and \( x_4 \) are concatenated along the channel dimension to form a comprehensive feature representation:
\begin{equation}\label{eq:local_feature_concatenation}
    x_{\text{concat}} = \mathrm{Concat}(x_1, x_2, x_3, x_4) \in \mathbb{R}^{512 \times N},
\end{equation}
which is further processed by a 1D convolution to produce a compact embedding $x_{\text{emb}} = \mathrm{Conv}(x_{\text{concat}}) \in \mathbb{R}^{\text{emb\_dims} \times N}$
Global descriptors are obtained by applying both adaptive max pooling and adaptive average pooling along the spatial dimension: $x_{\text{max}} = \mathrm{AdaptiveMaxPool1D}(x_{\text{emb}})$ and $x_{\text{avg}} = \mathrm{AdaptiveAvgPool1D}(x_{\text{emb}})$
which are concatenated to form a vector of dimension \( 2 \times \text{emb\_dims} \): $x_{\text{global}} = \mathrm{Concat}(x_{\text{max}}, x_{\text{avg}}) \in \mathbb{R}^{2 \times \text{emb\_dims}}$
\subsubsection{Stage 5: Classification}
The global feature vector is then finally passed through a series of fully connected layers. The final classification is produced by a linear layer mapping the feature vector to the desired number of output channels.
\section{EXPERIMENTAL RESULTS}
In this section, we assess the accuracy of the proposed architecture. We conducted ablation studies and compared the model's performance with state-of-the-art methods on two datasets.
\subsection{Datasets}
\subsubsection{MMActivity Dataset}
The MMActivity dataset~\cite{RadHAR} is a point cloud-based human activity dataset featuring five activities performed by two subjects, collected using TI’s IWR1443BOOST sensor board. It contains 93 minutes of data, with 71.6 minutes for training and 21.4 minutes for testing. To capture temporal dependencies, 2-second windows (60 frames) were created with a 10-frame sliding window, yielding 12,097 training samples and 3,538 test samples. Additionally, 20\% of the training data was used for validation.
\subsubsection{MiliPoint Dataset}
The MiliPoint dataset~\cite{Milipoint} contains point cloud data for human activity recognition and skeleton keypoint data for pose estimation. In this study, we used the point cloud data, comprising 49 activities performed by 11 subjects, collected with the TI IWR1843 mmWave radar. The dataset was divided into 80\% for training, 10\% for validation, and 10\% for testing.
\subsection{Implementation Details}
The model was trained using the Stochastic Gradient Descent (SGD) optimiser with a cosine annealing learning rate scheduler to adjust the learning rate throughout training. An early stopping mechanism monitored the validation loss to prevent overfitting, and model checkpointing saved the best-performing model on the validation set. The cross-entropy loss function was employed. Although the initial learning rate argument was set to 0.001, the effective initial learning rate for SGD was scaled to 0.1, with a momentum of 0.9 and a weight decay of 0.0001. A batch size of 32 was used during training.
\subsection{Ablation Study}
We conducted an ablation study to assess the model components and development process. For all evaluations, we set the number of heads to 1, $K$ to 20, and used four layers. First, we trained the model using only the MAK module, feeding its output directly to the classification module. Initially, feature fusion was excluded and achieved 92.40\% accuracy. When feature fusion was introduced, accuracy improved to 96.82\%, demonstrating the benefits of combining lower- and higher-level features. Next, we incorporated standard graph convolution, alternating it with the MAK layer (sandwich approach). This reduced Multiply-Accumulate operations (MACs) and parameters but slightly lowered accuracy to 96.76\%. Finally, we tested a sequential approach, with MAK layers first and graph convolution layers last. This further reduced MACs and parameters while increasing accuracy to 97.45\%, making it the optimal configuration. The improved performance of the sequential approach suggests that extracting local features first before applying global feature aggregation leads to better feature representation and classification accuracy. Results are summarised in Table 4.
\begin{table}[ht]
\caption{Evaluation of the model components. FF: Feature fusion, GC: Graph Convolution, SW: Sandwich, Sq: Sequential}
\label{MMAtivity_ablation_study}
\begin{center}
\resizebox{0.45\textwidth}{!}{
\begin{tabular}{|c|c|c|c|}
\hline
Method & MACs (G) & Params. (M) & Accuracy (\%)\\  
\hline
MAK & 16.43 & \textbf{1.59} & 92.40 \\
MAK+FF & 16.72 & 2.44 & 96.82 \\
MAK+FF+GC (Sw) & 5.87 & 1.95 & 96.76 \\
MAK+FF+GC (Sq)  & \textbf{3.95} & 1.86 & \textbf{97.45} \\
\hline
\end{tabular}
}
\end{center}
\end{table}
\subsection{Hyper Parameter Tunning}
To analyse parameter selection for the model architecture, we conduct experiment on both datasets.
\subsubsection{Effect of number of heads on accuracy and computational cost}w
To enhance the MHDF module’s feature extraction capability, we introduced a multi-head kernel to capture diverse local features. We conducted experiments to determine the optimal number of heads, as shown in Tables~\ref{MMAtivity_No_head_effect} and \ref{Milipoint_No_head_effect}. As a baseline, we set the K nearest neighbour to 20. 
The results indicate that increasing the number of heads raises the computational cost, as reflected in the higher MACs and model parameters. However, accuracy does not consistently improve with more heads. For the MMActivity dataset (Tab.~\ref{MMAtivity_No_head_effect}), the MAK with a single head achieved the highest accuracy (97.45\%) with the lowest computational cost. For the MiliPoint dataset (Tab.~\ref{Milipoint_No_head_effect}), the MAK with five heads attained the highest accuracy (99.12\%), though its computational cost was not the lowest.
Thus, the optimal number of heads depends on the dataset and application-specific requirements, such as computational efficiency.
\begin{table}[ht]
\caption{Effect of number of heads on accuracy and computational cost using MMAtivity dataset: Best result is highlighted bold}
\label{MMAtivity_No_head_effect}
\begin{center}
\resizebox{0.4\textwidth}{!}{
\begin{tabular}{|c|c|c|c|}
\hline
No. Heads & MACs (G) & Params. (M) & Accuracy (\%)\\  
\hline
1 & \textbf{3.95} & \textbf{1.86} & \textbf{97.45} \\
2 & 5.05 & 1.91 & 97.28 \\
3 & 6.15 & 1.96 & 96.74 \\
4 & 7.24 & 2.01 & 96.67 \\
5 & 8.34 & 2.06 & 96.83 \\
6 & 9.44 & 2.11 & 97.19 \\
7 & 10.54 & 2.15 & 96.23 \\
\hline
\end{tabular}
}
\end{center}
\end{table}
\begin{table}[ht]
\caption{Effect of number of heads on accuracy and computational cost using Milipoint dataset: Best result is highlighted bold}
\label{Milipoint_No_head_effect}
\begin{center}
\resizebox{0.4\textwidth}{!}{
\begin{tabular}{|c|c|c|c|}
\hline
No. Heads & MACs (G) & Params. (M) & Accuracy (\%)\\  
\hline
1 & \textbf{3.95} & \textbf{1.87} & 97.85 \\
2 & 5.05 & 1.92 & 98.02 \\
3 & 6.15 & 1.97 & 99.07 \\
4 & 7.24 & 2.02 & 98.87 \\
5 & 8.34 & 2.07 & \textbf{99.12}\\
6 & 9.44 & 2.12 & 98.99 \\
7 & 10.54 & 2.17 & 99.01 \\
\hline
\end{tabular}
}
\end{center}
\end{table}
\subsubsection{Effect of number of neighbours (K) on accuracy and computational cost}
We conducted an experiment to determine the optimal number of K nearest neighbours. Various values of K were tested, and the results are presented in Tab.~\ref{MMAtivity_No_k_effect} and \ref{MiliPoint_No_k_effect}. Since 1 and 5 heads yielded the highest accuracy for MMActivity and MiliPoint dataset respectively, we set the number of heads to 1 and 5. The results show that while MACs increase with K—where the smallest K had the lowest MACs—the number of model parameters remains unchanged. In terms of accuracy, the highest values 97.54\% and 99.28\% for the MMActivity and MiliPoint datasets were achieved when K was equal to 30.
\begin{table}[ht]
\caption{Effect of number of neighbours (K) on accuracy and computational cost using MMAtivity dataset: Best result is highlighted bold}
\label{MMAtivity_No_k_effect}
\begin{center}
\resizebox{0.4\textwidth}{!}{
\begin{tabular}{|c|c|c|c|}
\hline
No. K & MACs (G) & Params. (M) & Accuracy (\%)\\  
\hline
5 & \textbf{1.43} & 1.86 & 93.95 \\
10 & 2.27 & 1.86 & 97.31 \\
15 & 3.11 & 1.86 & 97.04 \\
20 & 3.95 & 1.86 & 97.45 \\
25 & 4.79 & 1.86 & 96.43 \\
30 & 5.63 & 1.86 & \textbf{97.54} \\
35 & 6.47 & 1.86 & 96.54 \\
40 & 7.31 & 1.86 & 95.30 \\
\hline
\end{tabular}
}
\end{center}
\end{table}
\begin{table}[ht]
\caption{Effect of number of neighbours (K) on accuracy and computational cost using MiliPoint dataset: Best result is highlighted bold}
\label{MiliPoint_No_k_effect}
\begin{center}
\resizebox{0.4\textwidth}{!}{
\begin{tabular}{|c|c|c|c|}
\hline
No. K & MACs (G) & Params. (M) & Accuracy (\%)\\  
\hline
5 & \textbf{2.53} & 2.07 & 98.60 \\
10 & 4.47 & 2.07 & 98.86 \\
15 & 6.40 & 2.07 & 98.89 \\
20 & 8.34 & 2.07 & 99.12 \\
25 & 10.28 & 2.07 & 99.12 \\
30 & 12.22 & 2.07 & \textbf{99.28} \\
35 & 14.44 & 2.07 & 98.26 \\
40 & 16.10 & 2.07 & 97.55 \\
\hline
\end{tabular}
}
\end{center}
\end{table}
\subsection{Comparison to State-of-the-Art Methods}
We compare the accuracy of the proposed model architecture with state-of-the-art methods using the MMActivity~\cite{RadHAR} and MiliPoint~\cite{Milipoint} datasets, with results presented in Tables~\ref{MMActivity_comparison} and~\ref{Milipoint_comparison}.
To ensure a fair comparison, both datasets were divided into training, validation, and test sets following the approach in~\cite{Milipoint,RadHAR}. For MMActivity, 12,097 samples were used for training, with 20\% for validation, and 3,538 samples for testing. For MiliPoint, the dataset was split into 80\% for training, 10\% for validation, and 10\% for testing and the training was done three times with different random seeds and the average was then computed. 
Our model outperforms existing methods, achieving the highest accuracy of 97.54\% on MMActivity (Tab.~\ref{MMActivity_comparison}) and 98.25\% on MiliPoint (Tab.~\ref{Milipoint_comparison}). These results highlight the effectiveness of the proposed model for action recognition.
\begin{table}[ht]
\caption{Accuracy Comparison on MMActivity Dataset: Best result is highlighted bold}
\label{MMActivity_comparison}
\begin{center}
\resizebox{0.4\textwidth}{!}{
\begin{tabular}{|c|c|c|c|c|}
\hline
Method & Acc. (\%) & Pre. & Rec. & F1\\  
\hline
SVM~\cite{RadHAR} & 63.74 & - & - & - \\
MLP~\cite{RadHAR} & 80.34 & - & - & - \\
BiLSTM~\cite{RadHAR} & 88.42 & - & - & - \\
TD-CNN-BiLSTM~\cite{RadHAR} & 90.47 & - & - & - \\ 
LPN-GRU~\cite{RobHAR} & 94.05 & 96.60 & 94.10 & 94.21 \\ 
LPN-BiLiLSTM~\cite{RobHAR} & 95.12 & 95.85 & 95.18 & 95.29 \\
ST-GCN~\cite{Gawon} & 96.55 & - & - & - \\
MAK-GCN~(Ours) & \textbf{97.54} & \textbf{97.58} & \textbf{97.54} & \textbf{97.54} \\
\hline
\end{tabular}
}
\end{center}
\end{table}
\begin{table}[ht]
\caption{Accuracy Comparison on Milipoint Dataset~\cite{Milipoint}: Best result is highlighted bold}
\label{Milipoint_comparison}
\begin{center}
\resizebox{0.3\textwidth}{!}{
\begin{tabular}{|c|c|}
\hline
Method & Accuracy (\%)\\  
\hline
DGCNN~\cite{Milipoint} & 13.61 \\   
Pointformer~\cite{Milipoint} & 29.27 \\ 
PointNet++~\cite{Milipoint} & 34.45 \\ 
PointMLP~\cite{Milipoint} & 18.37 \\ 
MAK-GCN~(Ours) & \textbf{98.25}\\
\hline
\end{tabular}
}
\end{center}
\end{table}
\begin{figure}[thpb]
  \centering
  \includegraphics[scale=0.3]{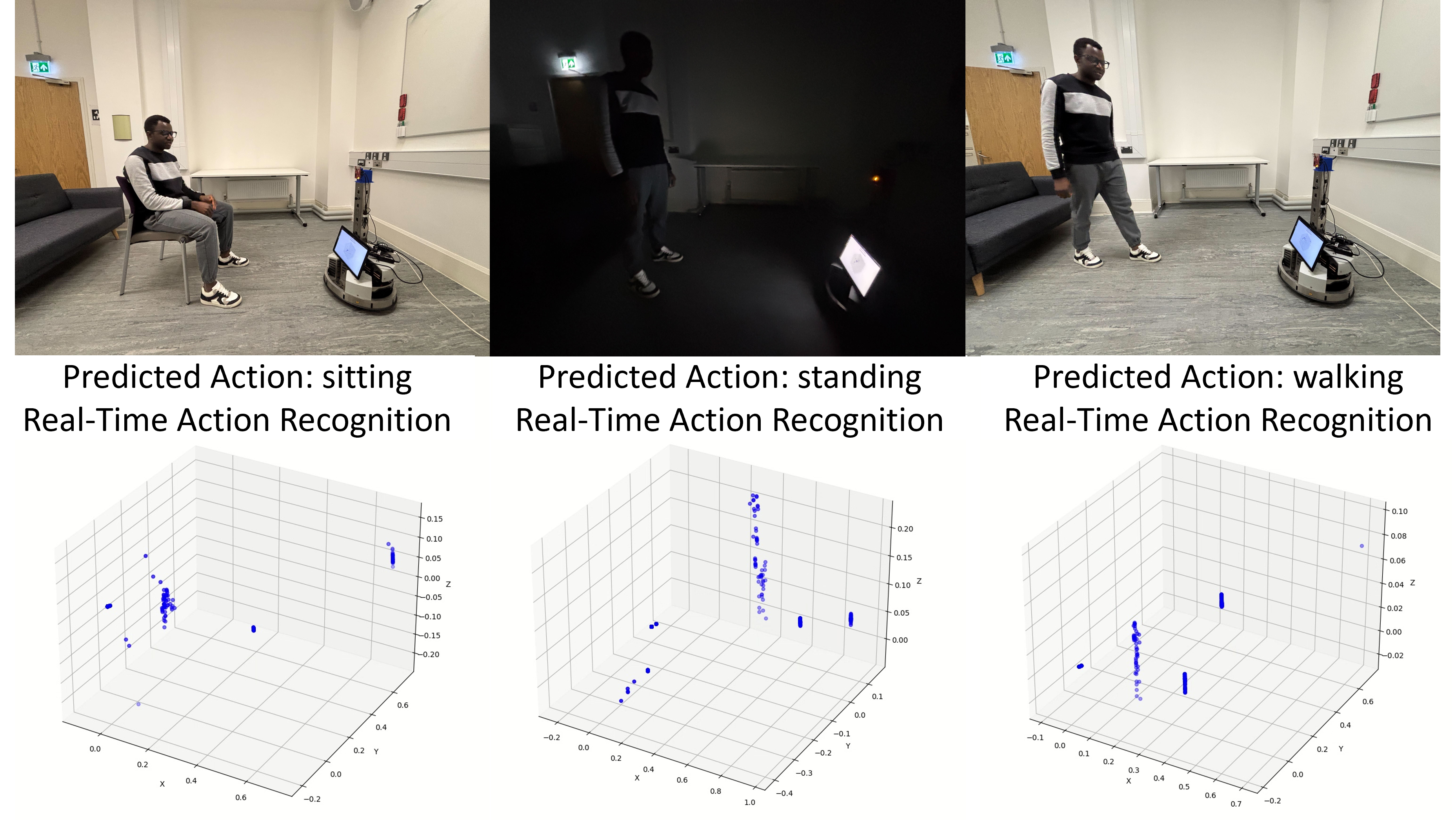}
  \caption{Human activity recognition with Robotino robot}
  \label{fig:demo}
\end{figure}
\section{APPLICATION FOR HUMAN ACTIVITY RECOGNITION}
To evaluate the performance of the proposed model, we tested it using a Robotino robot for human activity recognition. Data was acquired using the AWR1843BOOST mmWave radar, mounted on the robot, and transmitted via USB to a Jetson Nano using ROS (Robot Operating System) messages\footnote{\url{https://github.com/Gbouna/mmwave_data_collector}.}. The received messages were converted into point cloud data for activity recognition. To maintain consistency across frames, we set an upper limit $K$ on the number of points per frame. Frames exceeding $K$ were randomly sampled, while those with fewer points were zero-padded. To capture temporal dependencies, we stacked 50 frames (2 seconds). The trained model and action recognition pipeline were deployed on the Jetson Nano. Incoming data was processed and fed into the model for real-time recognition. As shown in Fig.~\ref{fig:demo}, the system successfully recognised actions under both lit and dark conditions, demonstrating its potential for real-world human activity monitoring applications.
\section{CONCLUSIONS}
To enhance the activity-monitoring capabilities of robots in home environments, this research proposes a novel model architecture that progressively refines local features through a multi-head adaptive kernel. The proposed model dynamically tailors convolution kernels to diverse feature correspondences, thereby enabling robust and discriminative feature learning for human activity recognition. Our proposed method pushes the boundaries of state-of-the-art by a significant margin upon two challenging datasets. With over $90\%$ accuracy, our proposed method shows potentials of realizing mmWave-based activity recognition solutions in real-world applications to address user acceptance issues and ultimately improving elderly people's quality of life. In future work, we intend to deploy our system in the homes of older people to assess its applicability in real-world settings.









\bibliographystyle{unsrt}
\bibliography{ref}

\end{document}